\documentclass[letterpaper, 10 pt, conference]{ieeeconf}  

\IEEEoverridecommandlockouts                              

\usepackage{graphicx}      
\usepackage{algorithm} 
\usepackage{algpseudocode} 
\usepackage{amsfonts}
\usepackage{amsmath}
\usepackage{url}
\usepackage{bbm} 

\usepackage[caption=false,font=footnotesize]{subfig}
\newcommand{\argmin}{\mathop{\mathrm{argmin}}}  
  
\title{\LARGE \bf
Grasp Planning for Customized Grippers by Iterative Surface Fitting
}
\author{Yongxiang Fan$^{*}$, Hsien-Chung Lin$^{*}$, Te Tang, Masayoshi Tomizuka
\thanks{$^*$These authors equally contributed to this work.}
\thanks{Yongxiang Fan, Hsien-Chung Lin, Te Tang, and Masayoshi Tomizuka are with Department of Mechanical Engineering, 
        University of California, Berkeley, Berkeley, CA 94720, USA
        {\tt\small {yongxiang\_fan, hclin, tetang, tomizuka}@berkeley.edu}}%
}

\begin{document}
\maketitle
\thispagestyle{empty}
\pagestyle{empty}

\begin{abstract}
Customized grippers have broad applications in industrial assembly lines. Compared with general parallel grippers, the customized grippers have specifically designed fingers to increase the contact area with the workpieces and improve the grasp robustness. However, grasp planning for customized grippers is challenging due to the object variations, surface contacts and structural constraints of the grippers. In this paper, an iterative surface fitting (ISF) algorithm is proposed to plan grasps for customized grippers. ISF simultaneously searches for optimal gripper transformation and finger displacement by minimizing the surface fitting error. A guided sampling is introduced to avoid ISF getting stuck in local optima and improve the collision avoidance performance. The proposed algorithm is able to consider the structural constraints of the gripper and plan optimal grasps in real-time. The effectiveness of the algorithm is verified by both simulations and experiments. The experimental videos are available at~\cite{website}. 
\end{abstract}

\section{Introduction}
Grasping is an essential capability for robots to extend the functionality and execute complex tasks such as assembly, picking and packaging. Compared with general parallel grippers, the customized grippers are designed for particular class of manipulation tasks by matching the contact surface of the gripper with the geometry of the workpieces. As a result, the customized grippers generally have larger gripping force and more robust grasp. For example, the food industry may design particular grippers to grasp cakes from conveyors for food packing, and the autonomous assembly line may require customized grippers to assure stable grasp and precise localization.

Considering the specific surface matching problem in grasp planning, this paper proposes an iterative surface fitting (ISF) algorithm, where it fits the contact surface based on a metric measuring the distance of gripper-object surface as well as the misalignment of the contact normals. ISF takes the motions of both the palm and fingers into consideration by a proposed iterative palm-finger optimization (IPFO). A guided sampling is also introduced to encourage the exploration of the regions with high fitting scores. 

The contributions of this paper are as follows. First, the proposed algorithm achieves simultaneous surface fitting and gripper kinematic planning by considering both the desired grasp surfaces and the structural constraints of the gripper such as the jaw width and the degree of freedoms (DOFs). 
Second, the proposed guided sampling avoids getting stuck in the local optima by exploiting the previous grasping experience.   
The grasp planning by ISF and guided sampling achieves a real-time planning and the time to search for a collision-free grasp is less than 0.1 s in average for the objects in the simulation and experiment.  
Furthermore, by combining with the dedicated gripper design, the proposed surface fitting algorithm can deal with objects with complicated shapes and those in a heavy clutter environment with unsegmented point clouds. The experimental videos are available at~\cite{website}.

\section{Related Work}
The grasp planning for customized grippers is challenging. 
The traditional point contact model or soft finger model~\cite{murray1994mathematical} in grasp modeling is insufficient to describe the large surface contact between the gripper and the object. Thus, the quality evaluation is ambiguous for one particular grasp configuration. Therefore, it is desirable to exploit the surface information of the gripper and the object for grasp planning. 

There are several related works proposed to reveal the importance of contact surfaces in grasp planning. A taxonomy of grasp is constructed to analyze grasp models and design grippers~\cite{cutkosky1986modeling, cutkosky1989grasp}. The taxonomy observes that both power grasps and precision grasps tend to increase the contact surface to improve the stability and robustness of the grasp, where the object is either surrounded by the palm and fingers in power grasps, or by the soft fingertips in precision grasps.

In~\cite{hang2014hierarchical}, a fingertip space is proposed to take into account the matching of the basic fingertip geometry to object surface. The grasps are searched based on several layers of fingertip space with different resolutions. This method assumes point contact between the finger and the object, which is not the case for customized grippers with large contact surfaces. 
To take advantage of the large contact surface on grasp stability and robustness, the optimal grasps in~\cite{ciocarlie2007dexterous} are searched by minimizing the distance between predefined points on the hand and the surface of the object. The resultant grasps match the object surface by enclosing the object with more contacts. However, the computation load is excessively heavy for online implementations. 
The grasp synthesis for human hands using shape matching algorithm is proposed in~\cite{li2007data}. The algorithm matches hand shapes in a database to the query object by identifying collections of features with similar relative placements and surface normals. However, this approach requires lots of human demonstration to collect enough hand pose samples. Hence the searching for proper grasps  requires considerable time with exhaustive sampling. 

The idea of matching the shapes of parallel grippers to the geometries of objects is utilized in~\cite{klingbeil2011grasping,ten2018using} to accelerate the grasp searching speed by filling the parallel gripper with the object. However, this method cannot be generalized to a gripper with complicated shape like customized grippers.
In~\cite{levine2016learning, mahler2016dex}, the grasp strategy is trained in an end-to-end manner from image to grasp policy with millions of grasping data. The resultant policy reflects the importance of matching the fingertip with objects, while changing grippers requires re-collecting data and re-training the policy.  
 

In order to consider more complicated gripper shape, reduce the computation load, and retrieve reliable and secure grasps, it is desired to match the surfaces on the gripper to the object more precisely. The idea of shape registration using iterative closest point (ICP) is first proposed in~\cite{besl1992method} and then refined by~\cite{jost2003multi,low2004linear,zinsser2003refined}. 
However, these approaches only consider the alignment of one source object to one target object, while the surface matching in grasp planning for customized grippers can be regarded as registering multiple surfaces on different fingers to one target object. Moreover, the surfaces have to satisfy the constraints of the grippers such as the jaw width, allowable DOFs and the alignment of contact normals.


\section{Problem Statement}
\label{sec:problem_statement}
\begin{figure}[t]
	\begin{center}
	    {\includegraphics[width =0.8\linewidth]{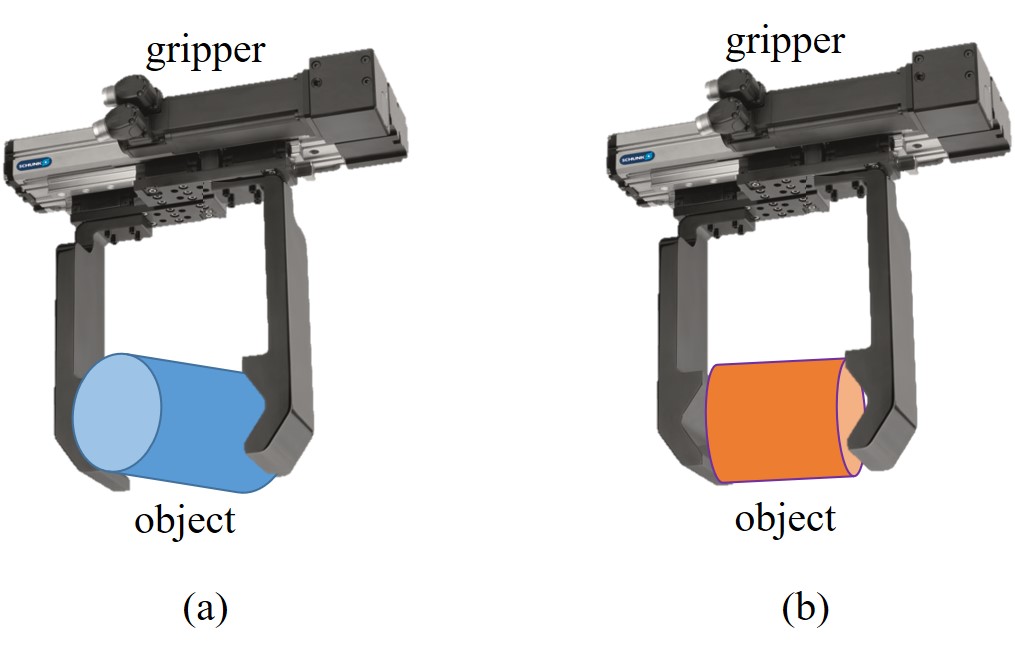}}
		\caption{An example grasp of cylinder objects by a parallel gripper with curved fingertip surfaces. A natural quality is the surface fitting error between the gripper and the object, under which the blue object in (a) has larger quality compared with the orange object in (b).}
		\label{fig:gripper_object}
	\end{center}
\end{figure}

A grasp planning example with a customized gripper is shown in Fig.~\ref{fig:gripper_object}. The customized gripper has parallel jaws with curved fingertip surfaces. The object to be grasped is a cylinder. The objective of the grasp planning is to search for the optimal pose of the gripper relative to the object by maximizing a quality metric considering the structural constraints of the gripper. More concretely, it is formulated as
\begin{subequations}
	\label{eq:general_form}
	\begin{align}
	\max_{R, t, \delta d, \mathcal{S}_j^f, \mathcal{S}_j^f} &\  Q(\mathcal{S}_1^f, \mathcal{S}_2^f,\mathcal{S}_1^o, \mathcal{S}_2^o) \label{eq1:cost}\\
	s.t. \quad 
	& \mathcal{S}_j^f \subset \mathcal{T}(\partial \mathcal{F}_j;R,t,\delta d), \quad j = 1,2 \label{eq1:surface_finger}\\
	& \mathcal{S}_j^o = NN_{\partial \mathcal{O}} (\mathcal{S}_j^f), \quad j = 1,2 \label{eq1:surface_object}\\
	& (\mathcal{S}_1^f, \mathcal{S}_2^f) \in \mathcal{W}(d_0 + \delta d) \label{eq1:constraint1}\\
	& d_0 + \delta d \in [d_{\text{min}}, d_{\text{max}}] \label{eq1:constraint2}
	\end{align}
\end{subequations}
where $j \in \{1,2\}$ is the finger index,  $R \in SO(3), t\in \mathbb{R}^3$ are the rotation and the translation of the gripper jaw from the original pose, $\delta d\in \mathbb{R}$ is the finger displacement from the original width $d_0$, and $Q$ represents the grasp quality with respect to the finger contact surfaces $\mathcal{S}_j^f$ and the object contact surface $\mathcal{S}_j^o$. The finger contact surface $\mathcal{S}_j^f$ lies on the finger surface $\partial \mathcal{F}_j$ transformed by $\mathcal{T}$, as shown in~(\ref{eq1:surface_finger}). The object contact surface $\mathcal{S}_i^o$ is determined by the nearest neighbor of the $\mathcal{S}_j^f$ on the object surface $\partial \mathcal{O}$, as shown in~(\ref{eq1:surface_object}).
Constraint~(\ref{eq1:constraint1}) indicates that the finger contact surfaces should be in the workspace $\mathcal{W}$ parameterized by the jaw width, and (\ref{eq1:constraint2}) describes the constraint of the finger displacement. The optimization~(\ref{eq:general_form}) searches for the optimal gripper transformation $(R^*,t^*)$ and finger displacement $d^*$ by maximizing the grasp quality $Q$.

Problem~\eqref{eq:general_form} is a standard grasp planning problem if the contact surfaces are degenerated into points. In general surface contact situation, the point contact model may not be able to incorporate the gripper surface directly into the planning. Problem~(\ref{eq:general_form}) becomes challenging to solve by either gradient based methods or sampling based methods. On one hand, modeling the contact surfaces as decision variables is nontrivial in the gradient based methods. On the other hand, the sampling based methods require exploring the whole state space, which is not applicable for real-time implementation. 

A natural surface-related quality can be constructed by matching the surfaces between the object and the gripper. Intuitively, the grasp with small surface matching error~(Fig.~\ref{fig:gripper_object}(a)) is more stable and robust compared with the one with large surface matching error~(Fig.~\ref{fig:gripper_object}(b)). Therefore, an iterative surface fitting (ISF) algorithm is proposed in this paper to search for optimal grasps of the gripper relative to the object. The optimality is in the sense of surface fitting errors. 
The formulation is modified from the iterative closest point (ICP) by including the structural constraints of different fingers such as the width and DOFs. ISF is initialized by a guided sampling algorithm in order to avoid being trapped in local optima and achieve collision avoidance. 

\section{Grasp Planning by Iterative Surface Fitting}
\label{sec:iterative_surface_fitting}
In this section, the related ICP is first reviewed in Section~\ref{sec:icp}, followed by the introduction of ISF with a one DOF customized gripper in Section~\ref{sec:isf}. The grasp planning with multi-fingered customized gripper is discussed in Section~\ref{sec:isf_multi}. Lastly, the guided sampling of ISF is presented in Section~\ref{sec:guided_sampling}. 

\subsection{Iterative Closest Point}
\label{sec:icp}
The ICP algorithm is first proposed in~\cite{besl1992method} for 3D shape registration. 
It searches for the optimal rigid transformation to align the source surface towards the target surface by minimizing the distance between them. 
The minimization is conducted by iteratively searching the correspondence points of the surfaces and minimizing the Euclidean distance of these correspondence points. More specifically, it is written as
\begin{equation} 
\label{eq2:icp_basic}
\begin{aligned}
 \min_{R,t}\  \sum_{i = 1}^{m}\|Rp_i + t - q_i\|_2^2,
\end{aligned}
\end{equation}
where $p_i, q_i \in \mathbb{R}^3$ denote a correspondence pair on the source surface and the target surface, and $m$ is the number of the correspondence pairs. The correspondence of $(p_i, q_i)$ are searched by the nearest neighbor method. The optimal transformation $(R,t)$ is calculated and applied to the source surface, after which the correspondence is updated by searching with the nearest neighbor again. These two steps are performed iteratively until convergence. 

A relevant variation calculates the distance along the normal direction~\cite{chen1992object}: 
\begin{equation} 
\label{eq2:icp_p2p}
\begin{aligned}
 \min_{R, t} \sum_{i = 1}^{m}\left((Rp_i + t - q_i)^Tn_{i}^q\right)^2,
\end{aligned}
\end{equation}
where $n_i^q$ indicates the normal of the target surface on point $q_i$. Instead of calculating the Euclidean distance between the transformed source point and the corresponding target point, the modification calculates the distance of the point to the plane by projecting the vector to normal direction $n_i^q$. Therefore, this modification allows the source surface to slide along the flat target surface.

With the small rotation angle assumption, the rotation matrix $R$ can be approximated by $I + \hat{r}$, where $\hat{r} \in so(3)$ represents the skew-symmetric matrix of the axis-angle vector $r\in \mathbb{R}^3$. 
With this approximation, (\ref{eq2:icp_p2p}) can be solved analytically by a standard least squares
\begin{equation} 
\label{eq2:icp_ls}
\begin{aligned}
 \min_{x} \| \mathbf{A} x - \mathbf{b}\|_2^2,
\end{aligned}
\end{equation}
where $\mathbf{A} = [a_1^T, ..., a_m^T]^T$, and $\mathbf{b} = [b_1, ..., b_m]^T$ with
\begin{subequations}
\label{eq2:ls_entry}
    \begin{align}
        a_i &= \left[ (p_i\times n_i^q)^T,\  (n_i^q)^T \right],\\
        b_i &= (q_i - p_i)^Tn_i^q.
    \end{align}
\end{subequations}
The optimal solution for the transformation is then given by $x =\left[r^T, \ t^T \right]^T = (\mathbf{A}^T\mathbf{A})^{-1}\mathbf{A}^T \mathbf{b}$ under the current correspondence. 


\begin{figure}[bt]
	\begin{center}
	    {\includegraphics[width =1\linewidth]{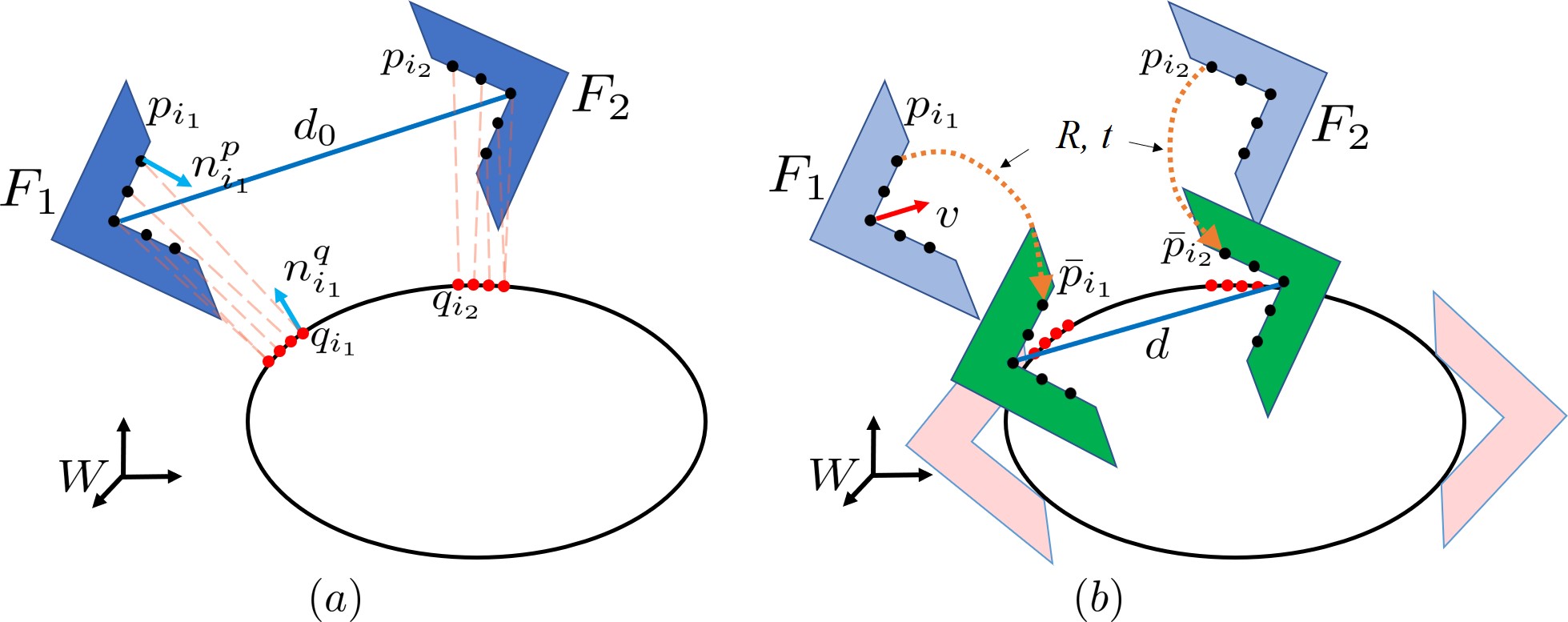}}
		\caption{Illustration of the iterative surface fitting (ISF). (a) The correspondence matching, where the corresponding point $q_{i_j}$ on object surface is found by the nearest neighbor searching and outliers/duplicates removal. (b) The surface fitting procedure, where the gripper transformation $(R,t)$ and finger displacement $\delta d$ are optimized by an iterative palm-finger optimization (IPFO). The green gripper is the updated result by the optimization, and the pink one is the converged result after several iterations of ISF.}
		\label{fig:isf}
	\end{center}
\end{figure}
ICP algorithm, however, can only register one single surface to the target surface at a time. In the grasping applications, a gripper usually consists of two or more fingers, which are all expected to be fitted on the target workpiece. We propose an iterative surface fitting (ISF) algorithm here to register multiple finger surfaces to the target workpiece at the same time considering the allowable DOFs of fingers.
\subsection{Iterative Surface Fitting with one DOF Gripper}
\label{sec:isf}
For ease of illustration, the ISF algorithm is introduced using a customized gripper with two fingers and one DOF, where the two fingers move in opposite directions to adjust the jaw width (Fig.~\ref{fig:isf}). The ISF algorithm iteratively executes two modules: the correspondence matching and the surface fitting. The correspondence matching contains a nearest neighbor search and an  outlier/duplicate filtering. As shown in Fig.~\ref{fig:isf}(a), given the point $p_{i_j}$ on the gripper surface (black dots) on the $j$-th finger, a KD-Tree is employed to search the nearest neighbor $q_{i_j}$ on the object surface (red dots), where the dash lines represent their correspondence.   
The outlier filtering removes the pairs whose distances are excessively large and the duplicate filtering considers the case where multiple points are assigned to the same point on the object. In this case, only the pair with the minimum distance will be kept~\cite{zinsser2003refined}. The outlier/duplicate filtering is particularly useful in our implementation since the gripper surface and object surface overlap only partially, and rejecting these point pairs greatly increases the robustness. 

In Fig.~\ref{fig:isf}(b), the surface fitting aims to search over the gripper transformation $(R,t)$ as well as the relative motion $\delta d = d - d_0$ between the fingers, where $d_0$ and $d$ indicate the jaw width before and during the optimization. 
To be more specific, the surface points are transformed to new locations by 
\begin{equation}
\label{eq3:p2pbar}
    \begin{aligned}
    & \bar{p}_{i_j} = Rp_{i_j} + t + \frac{1}{2}(-1)^j {Rv\delta d}, \quad j = 1,2
    \end{aligned}
\end{equation}
where $v\in \mathbb{R}^3$ is a unit vector pointing from $F_1$ to $F_2$, as shown by the red arrow in Fig.~\ref{fig:isf}(b). Then the matching of contact points can be quantified by an point matching error metric,
\begin{equation}
\label{eq3:Ep}
    \begin{aligned}
     E_p(R,t,\delta d) &  = \sum_{i=1}^{m} \sum_{j = 1}^{2}\left((\bar{p}_{i_j}-q_{i_j})^Tn_{i_j}^q\right)^2.
    \end{aligned}
\end{equation}
An additional normal alignment error metric is
\begin{equation}
\label{eq3:En}
    \begin{aligned}
     E_n(R)   = \sum_{i = 1}^m\left(\left(Rn_{i}^p\right)^Tn_{i}^q + 1 \right)^2,
    \end{aligned}
\end{equation}
where $E_n$ represents the misalignment error between the normal of the gripper $n_{i}^p$ and that of the object  $n_{i}^q$. By minimizing $E_n$, the normals of the finger surface are forced to align towards the normals of the object surface.

The surface fitting is to minimize the overall error
\begin{subequations}
	\label{eq3:overall}
	\begin{align}
	\min_{R, t, \delta d} &\  E(R,t,\delta d) \label{eq3:cost}\\
	s.t. \quad 
	& \delta d + d_0 \in [d_\text{min}, d_\text{max}] \label{eq3:surface_finger}
	\end{align}
\end{subequations}
where $E(R,t,\delta d) = E_p(R,t, \delta d) + \alpha^2 E_n(R)$ represents the surface fitting error and $\alpha \in \mathbb{R}$ is the weighting factor that balances the the point matching and the normal alignment. 

Problem (\ref{eq3:overall}) is nonlinear due to the coupling between the $R$ and $\delta d$. 
In this paper, the palm transformation $R, t$ and the finger displacement $\delta d$ are solved by an iterative palm-finger optimization (IPFO). The palm optimization optimizes for the optimal palm transformation $R^*, t^*$ with fixed $\delta d$, while the finger optimization optimizes for the optimal finger displacement $\delta d^*$ with fixed ($R, t$). 

The palm optimization can be formulated as a least squares problem that is similar to \eqref{eq2:icp_ls} with an augmented matrix $\tilde{\mathbf{A}} = [\mathbf{A}_1^T, \mathbf{A}_2^T, \mathbf{A}_n^T]^T$ and an augmented vector $\tilde{\mathbf{b}} = [\mathbf{b}_1^T, \mathbf{b}_2^T, \mathbf{b}_n^T]^T$, where $\mathbf{A}_j$ and $\mathbf{b}_j$ are the point matching of each finger surface modified from~\eqref{eq2:icp_ls}, with $p_i$ in~\eqref{eq2:ls_entry}  replaced by $\tilde{p}_{i_j} = p_{i_j} + 0.5(-1)^j v\delta d$ for $j = 1,2$ to consider the displacement of the finger. 
$\mathbf{A}_n$ and $\mathbf{b}_n$ are additional terms to align the contact normals. Derived from~\eqref{eq3:En}, we can get $\mathbf{A}_n = [ a_{n,1}^T, \cdots, a_{n,m}^T]^T$ and $\mathbf{b}_n = [ b_{n,1}, \cdots, b_{n,m}]^T$ with $a_{n,i} = [\alpha (n_i^p \times n_i^q)^T, 0_3^T]$ and $b_{n,i} = -\alpha (n_i^p)^Tn_i^q - \alpha$.
Hence, the palm optimization has the closed form and can be represented as
\begin{equation}
    \label{eq3:closed_form}
    x = (\tilde{\mathbf{A}}^T\tilde{\mathbf{A}})^{-1}\tilde{\mathbf{A}}^T\tilde{\mathbf{b}}.
\end{equation}


The finger optimization with fixed $(R,t)$ in~(\ref{eq3:overall}) is a one-dimensional constrained quadratic programming, 
\begin{subequations}
	\label{eq4:overall}
	\begin{align}
	\min_{\delta d} \ & \sum_{i=1}^m \sum_{j = 1}^{2} (b_{i_j} - a_{i_j}\delta d)^2 \label{eq4:s}\\
	s.t. \quad 
	& \delta d + d_0 \in [d_\text{min}, d_\text{max}] \label{eq4:con}
	\end{align}
\end{subequations}
where $a_{i_j} = 0.5 (-1)^{j-1} (Rv)^Tn_{i_j}^q$, and $b_{i_j} = (Rp_{i_j} + t - q_{i_j})^Tn_{i_j}^q$. 
The optimal finger relative motion is given by
\begin{equation} 
    \label{eq3:closed_formd}
    \delta d^* = \begin{cases} 
    d_\text{min} - d_0, & \text{if } \delta \hat{d} + d_0 < d_\text{min} \\
    \delta \hat{d}, & \text{if } d_\text{min} \leq \delta \hat{d} + d_0\leq d_\text{max} ,\\
    d_\text{max} - d_0, & \text{if } \delta \hat{d}  + d_0 > d_\text{max}\\
    \end{cases}
\end{equation}
with
\begin{equation}
    \delta \hat{d} = \frac{\sum_{i=1}^m \sum_{j = 1}^{2}a_{i_j}b_{i_j}}{\sum_{i=1}^m \sum_{j = 1}^{2}a_{i_j}^2}.
\end{equation}

The procedure of IPFO is summarized in Alg.~\ref{alg:dual}. 
\begin{algorithm} [t]
	\caption{Iterative Palm-Finger Optimization (IPFO)}\label{alg:dual}
	\begin{algorithmic}[1]
		\State \textbf{Input:} $(p_{i},q_{i}),\ (n_{i}^p, n_{i}^q),\ d_0$\label{dual:input}
		\State \textbf{Init:} Initialize $\delta d^* = 0,\  R^* = I,\  t^* = 0_3,\  e_p = \infty$ \label{dual:init}
		\While {$e_p - E(R^*,t^*, \delta d^*) > \Delta e $} \label{dual:t}
		\State $e_p \leftarrow  E(R^*,t^*, \delta d^*)$
        \State $\{R^*, t^*\} \leftarrow \min_{R,t}E(R,t,\delta d^*)$ by (\ref{eq3:closed_form}) \label{dual:iter1}
        \State $\delta d^* \leftarrow \min_{\delta d} E(R^*,t^*,\delta d)$ by (\ref{eq3:closed_formd})\label{dual:iter2}
		\EndWhile
		\State \Return $\{R^*, t^*, \delta d^*, e_p\}$
	\end{algorithmic}
\end{algorithm}
Given the correspondence $(p_{i},q_{i})$ and the corresponding normals $(n_{i}^p, n_{i}^q)$ as inputs, IPFO minimizes the error metric~(\ref{eq3:cost}) in an iterative manner~(Line~\ref{dual:iter1}-\ref{dual:iter2}). The iteration stops when the error reduction is less than a threshold $\Delta e$~(Line~\ref{dual:t}).

Inspired by~\cite{jost2003multi}, ISF is optimized hierarchically by searching with a multi-resolution pyramid, as shown in~Alg.~\ref{alg:isf}. 
The inputs to ISF include the initial gripper state $R_c, t_c, d_0$, the surfaces $\partial \mathcal{O}$ and $\partial \mathcal{F}$, and the parameters for hierarchical searching~(Line~\ref{isf:input}). $L, I_0$ and $\epsilon_0$ denote the level number, maximum iteration and error bound for convergence, respectively.  The gripper surface $\partial F$ is first transformed to the specified initial state~(Line~\ref{isf:init}). In each level of the pyramid,  $\partial \mathcal{F}$ is downsampled adaptively to $\mathcal{S}^f$ with different resolutions (Line~\ref{isf:down}). The while loop iteratively searches correspondence and solves for desired gripper motion. The correspondence matching is conducted with the nearest neighbor search (Line~\ref{isf:nn}) and with the outlier/duplicate filtering (Line~\ref{isf:filter}). The desired palm transformation and finger displacement are optimized by IPFO~(Line~\ref{isf:IPFO}). 
Both the surface $\partial \mathcal{F}$ and its sample $\mathcal{S}^f$ are transformed by the optimized gripper motion (Line~\ref{isf:updateS}-\ref{isf:update}).  
The while loop is terminated if IPFO gives a similar transformation in adjacent iterations.

\begin{algorithm}[tb]
	\caption{Iterative Surface Fitting (ISF)}\label{alg:isf}
	\begin{algorithmic}[1]
		\State \textbf{Input:} Initial state $R_c,t_c, d_0$, $\partial \mathcal{O}$, $\partial \mathcal{F}$, $L, I_0, \epsilon_0$ \label{isf:input}
		\State \textbf{Init:} $\partial \mathcal{F} = \mathcal{T}(\partial \mathcal{F}; R_c, t_c, d_0)$ \label{isf:init}
		\For {$l = L-1, \cdots, 0$} \label{isf:paraymid}
		\State $\mathcal{S}^f\leftarrow \texttt{downsample}(\partial \mathcal{F}, 2^l)$,  $I_l = I_0/2^l$,  $\epsilon_l= 2^l\epsilon_0$\label{isf:down}
		\State $\mathcal{S}_{0}^f \leftarrow \mathcal{S}^f$, $e_{s} \leftarrow \infty$, $\eta \leftarrow 0$, $it \leftarrow 0$
		\While {$\eta \notin [1 - \epsilon_l, 1 + \epsilon_l] $ and $it\texttt{++} < I_l$}
		\State $e_{s,p} \leftarrow e_s$
		\State $\mathcal{S}^o\leftarrow NN_{\partial O}(\mathcal{S}^f)$  \label{isf:nn}
		\State $\{\mathcal{S}^f, \mathcal{S}^o\} \leftarrow \texttt{filter}(\mathcal{S}^f, \mathcal{S}^o)$ \label{isf:filter}
		\State $\{R^*, t^*, \delta d^*, error\} \leftarrow \mathbf{IPFO}(\mathcal{S}^f, \mathcal{S}^o, d_0)$ \label{isf:IPFO}
		\State $\mathcal{S}^f \leftarrow \mathcal{T}(\mathcal{S}^f; R^*, t^*, \delta d^*)$ \label{isf:updateS} 
		\State $\partial \mathcal{F} \leftarrow \mathcal{T}(\partial \mathcal{F}; R^*, t^*, \delta d^*)$ \label{isf:update}
		\State $d_0 \leftarrow d_0 + \delta d^*$
		\State $e_s \leftarrow \|\mathcal{S}^f - \mathcal{S}_{0}^f\|, \ \eta \leftarrow e_s/e_{s,p}$ 
		\EndWhile
		\EndFor \label{isf:paraymid2} 
		\State \Return $\{ error, \partial \mathcal{F}\}$
	\end{algorithmic}
\end{algorithm}

\subsection{Iterative Surface Fitting (ISF) in General Case}
\label{sec:isf_multi}
The ISF algorithm is able to extend to multi-fingered grippers. Similar as above, we deal with the coupling by iteratively solving for palm transformation and finger displacement. The difference is that the finger displacement for a $K_f$-finger case can be represented as $\delta \theta = \left[\delta \theta_1^T,\cdots, \delta \theta_{K_f}^T\right]^T$ instead of the change of jaw width $\delta d$. 
The surface fitting problem can be formulated as 
\begin{subequations}
	\label{eq4:general}
	\begin{align}
	\min_{R, t, \delta \theta } &\  \sum_{j = 1}^{K_f} E_j(R,t, \delta \theta_j) \label{eq4:cost}\\
	s.t. \quad 
	& \theta_{\text{min},j} \leq \delta \theta_j + \theta_{0,j}\leq \theta_{\text{max},j}, \label{eq4:constraint}
	\\ &  j = 1,\cdots,K_f \nonumber 
	\end{align}
\end{subequations}
where $E_j$ is the fitting error of the $j$-th finger, $\theta_{0,j}$ is the initial position of the $j$-th finger. Similarly to the one DOF case in~\eqref{eq3:overall}, the problem~(\ref{eq4:general}) can be decomposed to a series of palm optimization and finger optimization. The palm optimization optimizes for the optimal palm transformation $R^*,t^*$ with fixed $\delta \theta_j$, while the finger optimization solves for the optimal finger displacement $\delta \theta_j$ with fixed $(R, t)$. The derivation of IPFO for multi-fingered hands is similar with our previous work on finger splitting~\cite{fan2018realtime}.  Compared with one DOF case, the finger optimization for multi-fingered grippers with $\delta \theta$ has higher dimensions and requires more computation to solve numerically. 

\subsection{Initialization and Sampling}
\label{sec:guided_sampling}
The initialization of ISF is important since ISF converges to local optima. Multiple initialization is desired for the algorithm to explore different regions of the object, so as to avoid getting trapped in bad local optima and achieve better collision-avoiding solutions. 

In this paper, the object is firstly partitioned into $K$ clusters by k-means clustering. The center of each cluster is regarded as a candidate initial position of the gripper for ISF, and the initial orientation of the gripper is randomly sampled.  

We build an empirical model to guide the sampling among $K$ candidates. Similar to the multi-armed bandit model, we record the error of ISF evaluated in each cluster center and compute the average regret accordingly. The average regret of each cluster center is used to guide the succeeding sampling. The sampling process is summarized in Alg.~\ref{alg:sampling}. 
\begin{algorithm}[t]
	\caption{Grasp Planning Algorithm}\label{alg:sampling}
	\begin{algorithmic}[1]
		\State \textbf{Input: }  $\partial \mathcal{O}, \partial \mathcal{F}$, center\# $K$, sample\# $K_s$, $d_0$ \label{alg:samplinginput} 
		\State \textbf{Init:} $C\leftarrow\texttt{k-means}(\partial \mathcal{O},K)$,$regret = 0_K$,$trial = 0_K$\label{alg:samplinginit}
		\For {$It = 1,\cdots, K_s$}
		\State Guided sampling:\par \label{alg:step1}
			\hskip\algorithmicindent $k^*\leftarrow \argmin_k regret(k)$\par
			\hskip\algorithmicindent $t_c = C(k^*), R_c\leftarrow \texttt{randRot}()$
		\State ISF evaluation:\par \label{alg:step2}
		\hskip\algorithmicindent $\{error, \partial \bar{\mathcal{F}}\} \leftarrow \mathbf{ISF}(R_c, t_c, d_0, \partial \mathcal{O}, \partial \mathcal{F})$\par
		\hskip\algorithmicindent $col  \leftarrow f_{\text{col}}(\partial\bar{\mathcal{F}}, \partial \mathcal{O})$
		\State Regret update: \par \label{alg:step3}
		\hskip\algorithmicindent $regret(k^*) \leftarrow \frac{regret(k^*)\cdot trial(k^*) + error}{trial(k^*) + 1}$\par
		\hskip\algorithmicindent $regret(k^*) \leftarrow (1 + \gamma\cdot col)regret(k^*)$ \par
		\hskip\algorithmicindent $trial(k^*)\leftarrow trial(k^*) + 1$
		\EndFor
	\end{algorithmic}
\end{algorithm}

The Alg.~\ref{alg:sampling} is fed by the surfaces $\partial \mathcal{O}, \partial \mathcal{F}$ and the parameters including the cluster number $K$, the total sampling times $K_s$ and initial jaw width $d_0$. 
The sample centers $C \in \mathbb{R}^K$ is generated by k-means clustering~(Line~\ref{alg:samplinginit}). The system stores $trial \in \mathbb{R}^K$ and $regret\in \mathbb{R}^K$ to represent the previous sampling and evaluation experience. 
The sampling is guided by the average regret, and the cluster center with the minimum regret is chosen as the initial position of the gripper for the following ISF, while the initial orientation $R_c$ is randomly sampled~(Line~\ref{alg:step1}). 
The gripper with the sampled initialization is evaluated by ISF and  collision check~(Line~\ref{alg:step2}). The $f_{\text{col}}(\cdot,\cdot)$ is a boolean function that returns 1 when the inputs have collision, i.e. two object surfaces have intersection. 
In~Line~\ref{alg:step3}, the average regret of the $k^*$-th sample is updated by considering the surface fitting error and the collision penalty. The $\gamma$ is a penalty factor that penalizes the average regret for the collided samples.

\section{Simulations and Experiments}
\label{res:sim_exp}
\begin{figure}[t]
	\begin{center}
		{\includegraphics[width =0.45\linewidth]{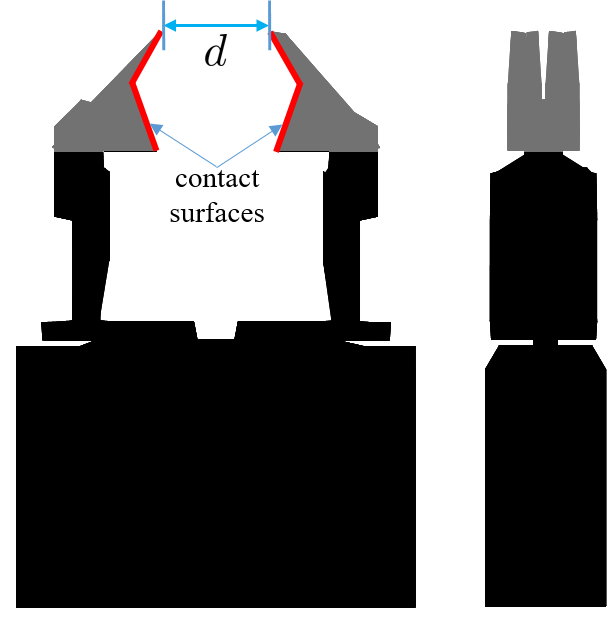}}
		\caption{The parallel jaw gripper used in simulations and experiments. The fingertips of the gripper and the contact surfaces are marked by gray and red colors, respectively. The allowed finger motion is shown by blue arrow, and the finger width $d$ is constrained within $[d_\text{min}, d_\text{max}]$. }
		\label{fig:gripper}
	\end{center}
\end{figure}

In this section, both the simulation and the experiment results are presented to verify the effectiveness of grasp planning algorithm.  The experimental videos are available at~\cite{website}. The computer we used was a desktop with 32GB RAM and 4.0GHz CPU. All the computations were conducted by Matlab. 
We used a SMC LEHF20K2-48-R36N3D parallel jaw gripper with the specialized fingers as shown in Fig.~\ref{fig:gripper}. The desired contact surfaces are marked by red. The gripper width was constrained by $\left[d_\text{min},d_\text{max} \right] = \left[1,3\right]$ cm.

\subsection{Parameter Lists}
The initial gripper width was set as $d_0$ = 2 cm. The weight for normal penalty was $\alpha = 0.01$. The convergence threshold was selected as $\Delta e = 10^{-5}$ in Alg.~\ref{alg:dual}. The pyramid level $L$, the maximum iteration $I_0$ and the tolerance $\epsilon_0$ were set as $4$, $200$ and $0.008$ respectively in Alg.~\ref{alg:isf}. The k-means center number $K = 6$, the sample number $K_s = 60$, and the collision penalty $\gamma = 0.2$ in Alg.~\ref{alg:sampling}.

\subsection{Simulations}
The grasp planning optimized for grasps by guided sampling and ISF evaluation. The proposed guided sampling enabled more efficient exploration by minimizing the average regret based on the previous experience. The proposed ISF algorithm considered the surface fitting by optimizing both the palm transformation and finger displacement. 

\begin{figure}[t]
	\begin{center}
		{\includegraphics[width =0.9\linewidth]{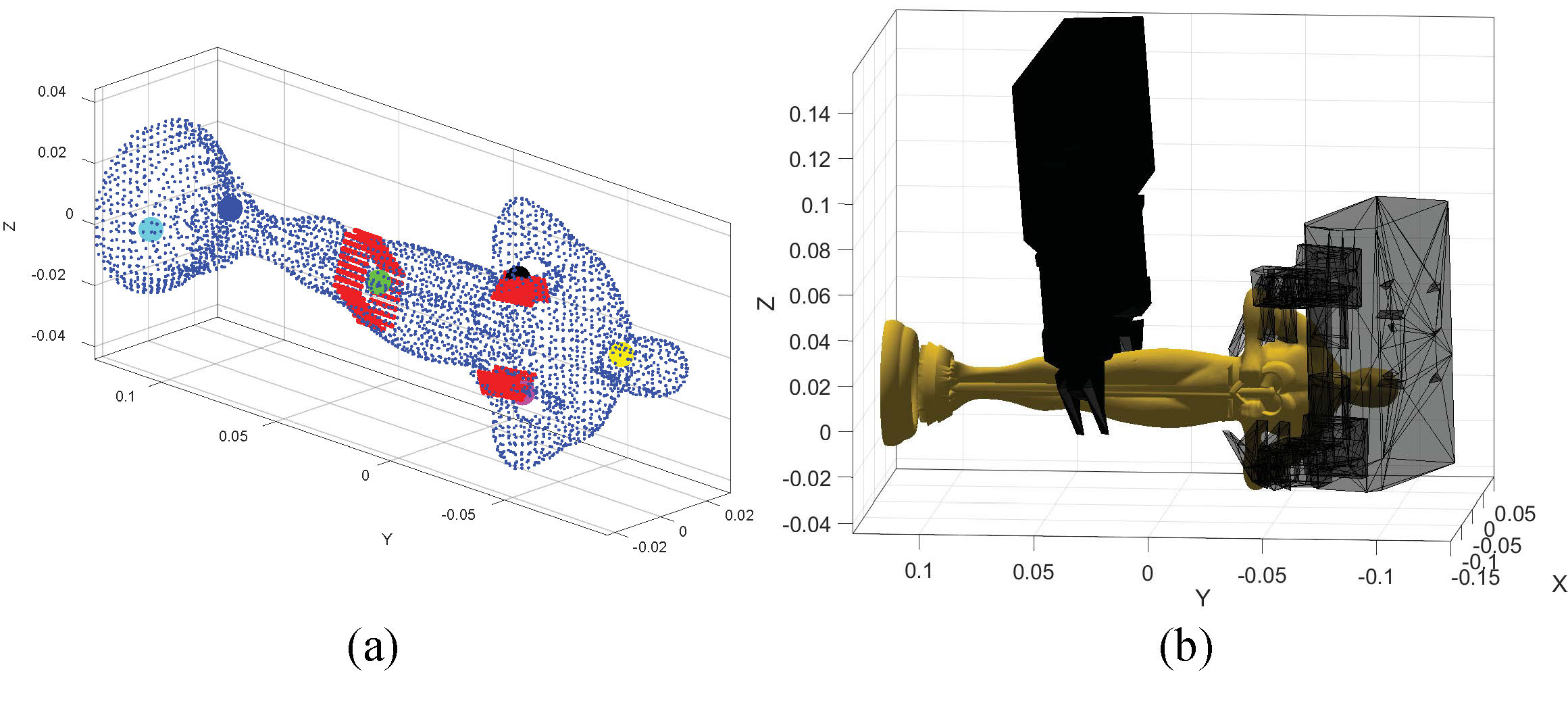}}
		\caption{ (a) Illustration of the grasp planning on an Oscar model, where the blue and red dots represent the object and the gripper surfaces, and the bold dots represent the centers of k-means. (b) The visualization of planned grasps, where the planned collision-free grasp is represented by the solid gripper, while the collided grasp is represented by the transparent one.}
		\label{fig:with_wo_collision}
	\end{center}
\end{figure}

Figure~\ref{fig:with_wo_collision} shows the grasp planning result on an Oscar model. The object and gripper surface are shown by the blue and red dots in Fig.~\ref{fig:with_wo_collision}(a). The vertices of the object were firstly processed by downsampling and normal estimation, after which the k-means clustering ran for centers of initialization, as shown by bold dots. 
Multiple grasps were generated (shown by red patches in~Fig.~\ref{fig:with_wo_collision}(a)) and passed through the collision check function. The planned collision-free grasp in the figure is represented by the pose of the solid gripper, while the collided grasp is represented by the pose of the transparent one as shown in Fig.~\ref{fig:with_wo_collision}(b).

\begin{figure}[t]
	\begin{center}
		{\includegraphics[width =1\linewidth]{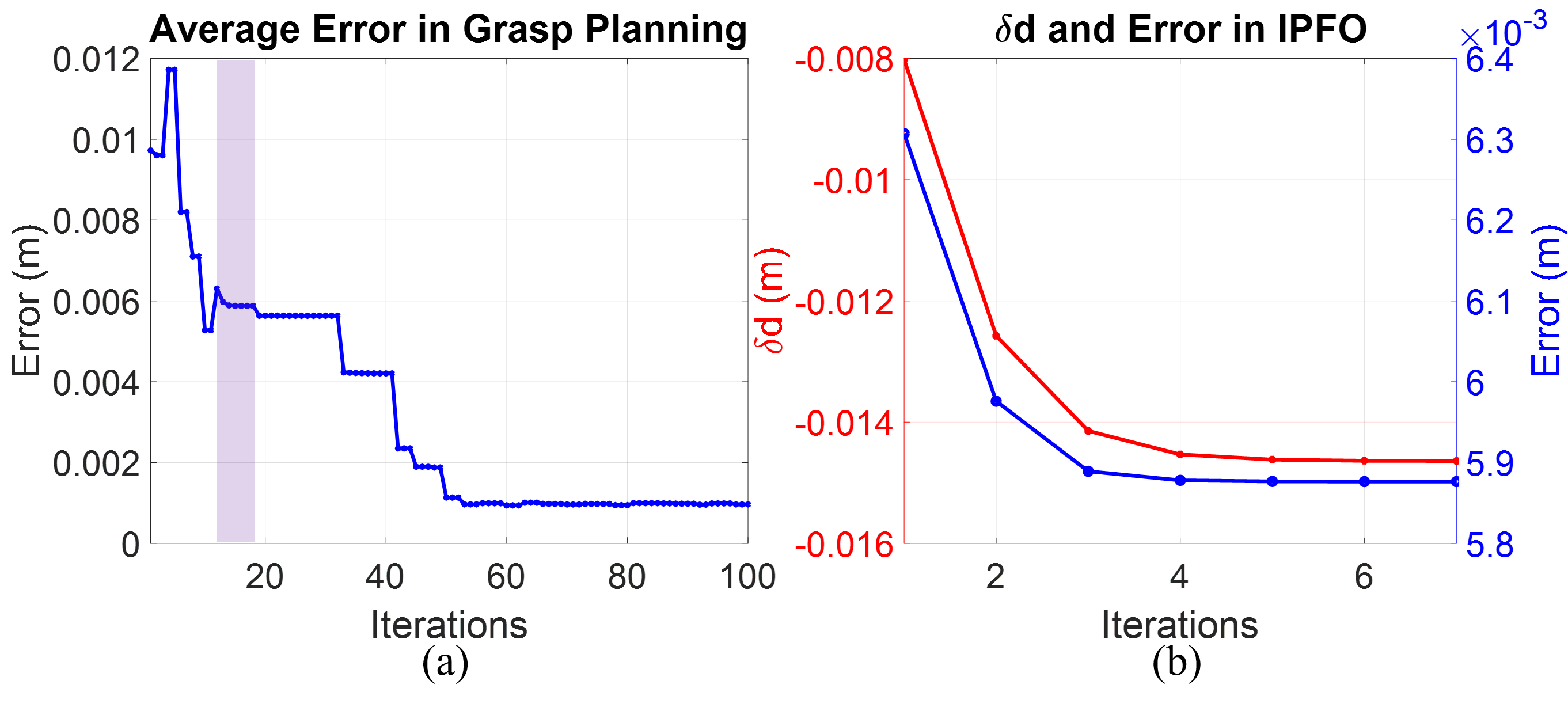}}
		\caption{Surface fitting errors of ISF and one execution of IPFO for grasp planning of Oscar model. (a) The error plot of the ISF algorithm. ISF is able to reduce the fitting errors from 10 mm to 1 mm by running IPFO despite certain oscillations caused by changing of the correspondence. (b) One execution of IPFO (shaded region in Fig.~\ref{fig:average_error_and_IPFO}(a)). The IPFO optimizes for palm transformation and finger displacement and converges in 7 iterations.}
		\label{fig:average_error_and_IPFO}
	\end{center}
\end{figure}

Figure~\ref{fig:average_error_and_IPFO} shows the surface fitting errors of ISF and one execution of IPFO during the grasp planning for the Oscar model. The average distance between $\partial \mathcal{F}$ and $\partial \mathcal{O}$ was dropped from 10 mm to 1 mm by running ISF as shown in Fig.~\ref{fig:average_error_and_IPFO}(a), where the shaded area is magnified and shown in Fig.~\ref{fig:average_error_and_IPFO}(b). The red line shows the finger displacement in different iterations and the blue line shows the fitting error. The IPFO could achieve efficient convergence within 7 iterations. 

\begin{figure}[t] 
	\begin{center}
		{\includegraphics[width =1\linewidth]{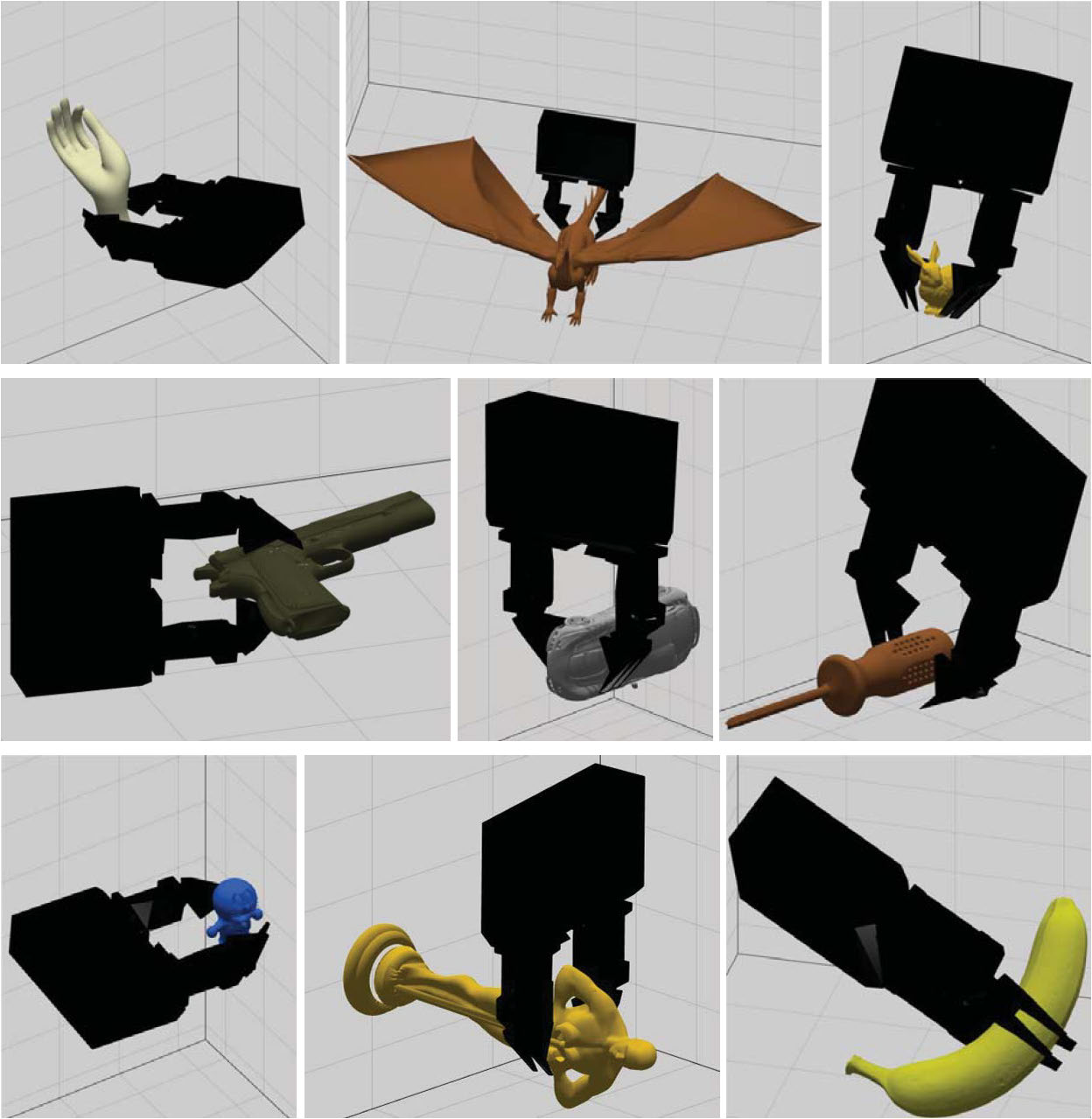}}
		\caption{Illustration of the grasp planning simulation on nine objects. }
		\label{fig:sim_res}
	\end{center}
\end{figure}

The simulation results on different objects are visualized in Fig.~\ref{fig:sim_res}. Some of the objects (e.g. Doraemon and Bunny) were scaled to fit into the gripper range. 
The simulation details for these objects are shown in Table~\ref{tab:opt_details}. The second column shows the number of collision-free optimal grasps over the total samples. The third column shows the total computation time (i.e. the time to run Alg.~\ref{alg:sampling}) to generate these grasps. The last column shows the number of vertices for each object. The dragon was challenging to grasp due to collision caused by the complex geometry.  In average, the grasp planning took 2.33 s to find 36.2 collision-free grasps in 60 times of samples. Thus, each collision-free grasp took 64.4 ms in average. 

\begin{table}[t]
\centering
\caption{Numerical Results of Grasp Planning Simulation}
\label{tab:opt_details}
\begin{tabular}{r|ccc}
Objects       & \begin{tabular}[c]{@{}c@{}} Collision Free Grasps\\ \hline Total Samples \end{tabular}  & \begin{tabular}[c]{@{}c@{}} $t_{\text{total}}$ \\ (sec.)\end{tabular} & \begin{tabular}[c]{@{}c@{}}Number of\\ Vertices\end{tabular} \\
\hline \hline
Hand         & 53/60  & 1.85  & 1939  \\
Gun          & 36/60  & 2.45  & 2287  \\
Car          & 18/60  & 1.68  & 2166  \\
Oscar        & 31/60  & 2.03  & 3684  \\
Dragon       & 3/60   & 5.61  & 3971  \\
Bunny        & 33/60  & 1.56  & 740   \\
Banana       & 54/60  & 3.82  & 1723  \\
Screw Driver & 58/60  & 1.71  & 930   \\
Doraemon     & 40/60  & 1.84  & 769   \\
\hline 
Average      & 36.2/60 & 2.33 & 2022.1
\end{tabular}
\end{table}

\subsection{Experiments}
A series of experiments were further performed on a FANUC LR Mate 200-iD/7L  industrial manipulator to verify the effectiveness of the proposed algorithm. Two IDS Ensenso N35 stereo camera sets were used to capture the point cloud of the object. Compared with simulation, the point cloud produced by Ensenso cameras was not able to reflect the object precisely due to occlusion and noise. The point cloud was smoothed and used to estimate the normals of the objects. 

\begin{figure}[t]  
	\centering
	{\includegraphics[width =1\linewidth]{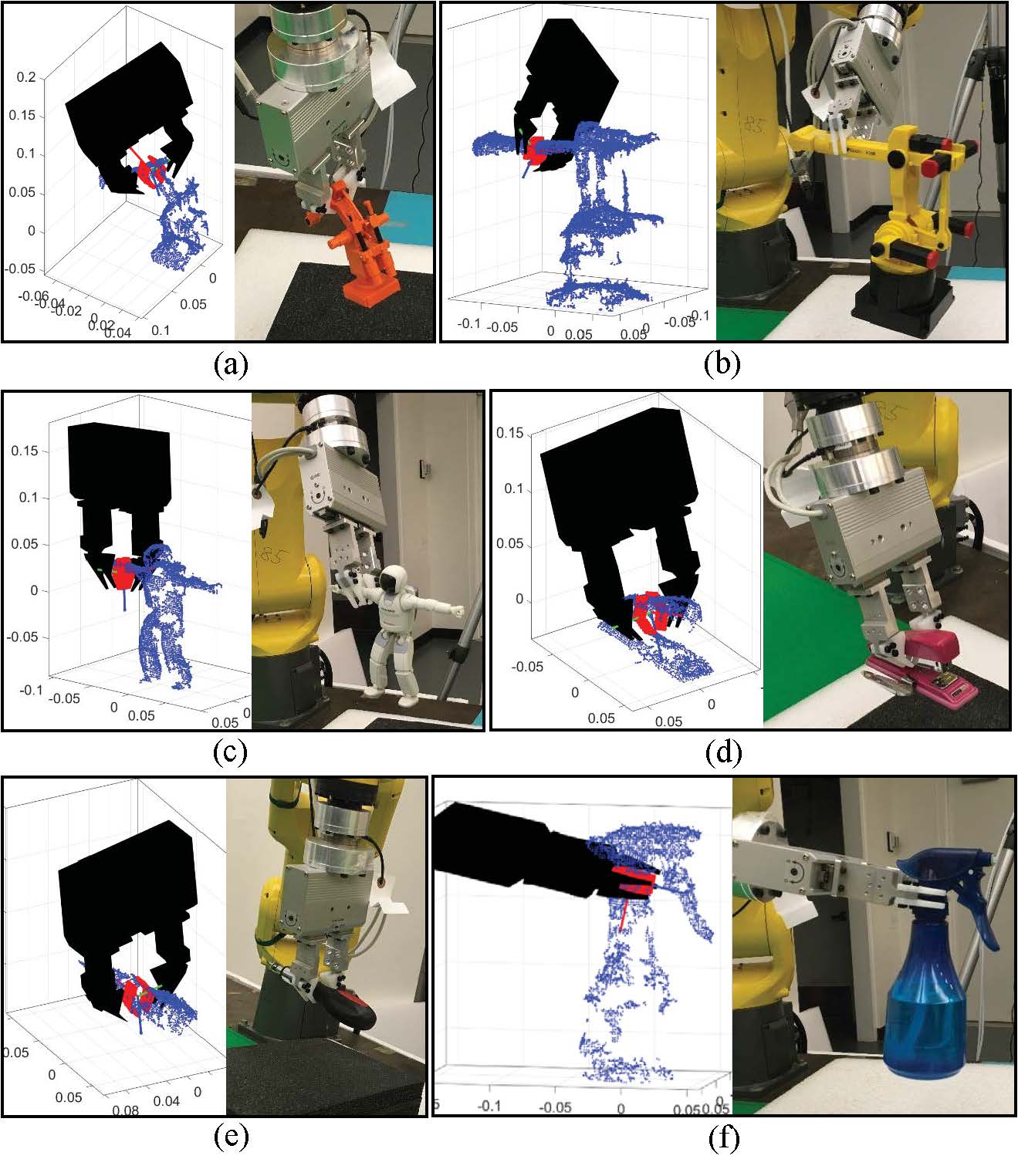}}
	\caption{The results of the grasp planning experiment on six objects.}
	\label{fig:graspExp1}
\end{figure}

Figure~\ref{fig:graspExp1} shows the grasp planning results on six different objects including three toy robot models and three tools in different sizes. The left side for each subfigure shows the observed point cloud and the optimized grasp, and the right side shows the grasp execution result in lifting the object by 10 cm. Although some of the objects had complicated shapes, ISF could find a grasp that matched the fingertips to the object surface well. Therefore, the robot could firmly grasp the object, and further increased the grasp robustness. 


\begin{figure}[t] 
	\centering
	{\includegraphics[width =0.95\linewidth]{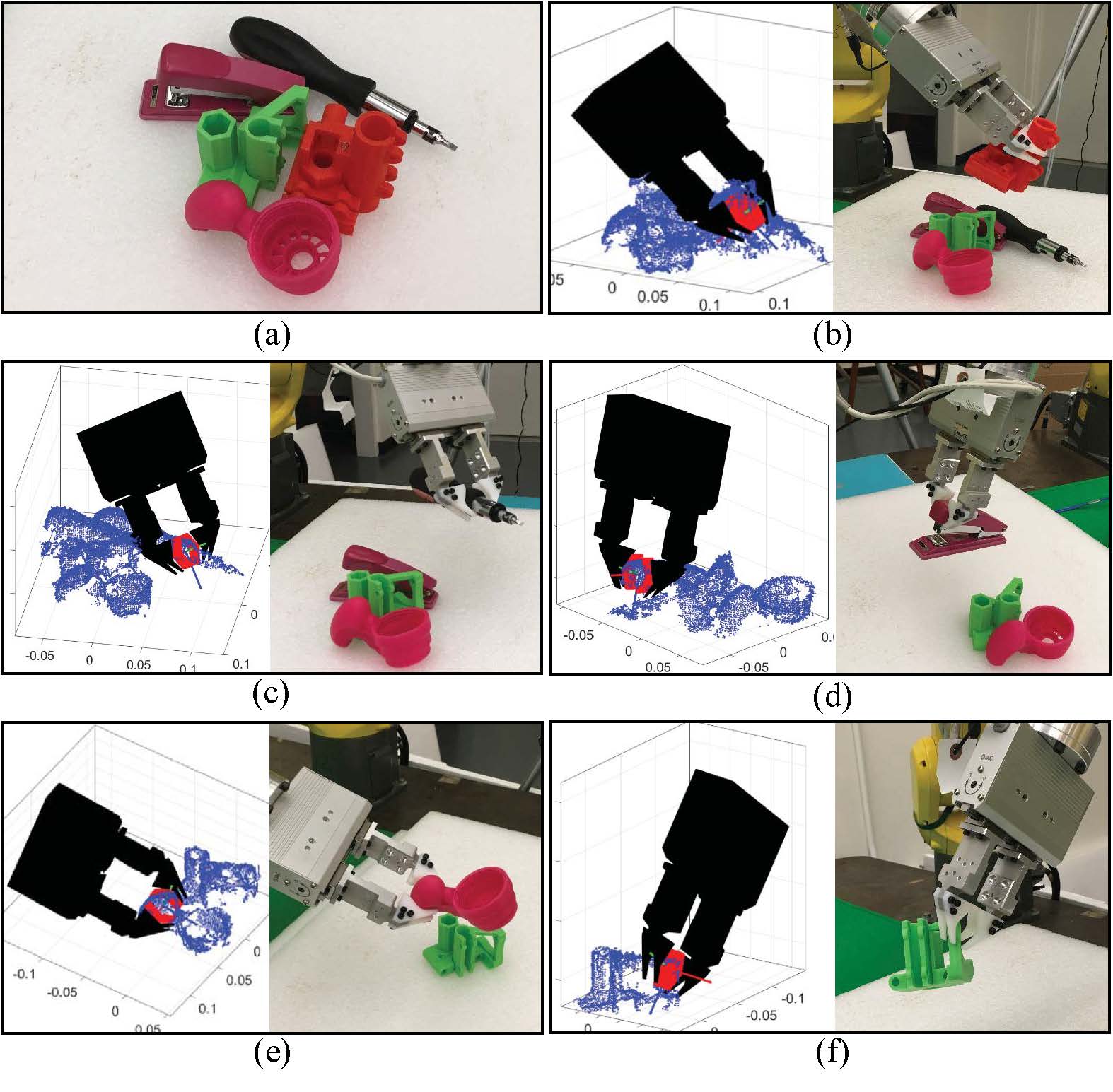}}
	\caption{The results of the grasp planning experiment in a heavy clutter environment (a) The initial object sets. (b)-(f) The consecutive grasps in the task.}
	\label{fig:graspExp2}
\end{figure}

A picking task in a heavy clutter environment was performed and shown in Fig.~\ref{fig:graspExp2}, where several objects were placed closely (Fig.~\ref{fig:graspExp2}(a)). Figure~\ref{fig:graspExp2}(b-f) show the consecutive grasps in the task. In this experiment, the proposed algorithm directly exploited grasp poses on the unsegmented point cloud by fitting the fingertip surface to the object sets. Even though the surface composed by the cluttered objects became more complicated than a single object, ISF filtered out the grasps with collision and found a suitable grasp to pick up objects sequentially. 


\section{Conclusions and Future Works} 
\label{sec:conclusion}
This paper proposed an iterative surface fitting (ISF) algorithm to plan grasps for customized grippers. ISF searched for optimal grasps by an iterative palm-finger optimization, which solved for the optimal palm pose and the finger displacement iteratively with closed-form solutions. To avoid bad local optima, the guided sampling was introduced to initialize ISF searching. The proposed grasp planning algorithm was applied to a series of simulations and experiments. ISF achieved 64.4 ms average searching time to find a collision-free grasp on the objects in simulations. The grasp planning was further implemented in a clutter environment to grasp objects from unsegmented point clouds. 

In the future, we would like to extend the algorithm to the grasp planning of general multi-fingered hands and soft hands. In addition, we would like to create an algorithm to learn the guided sampling by the grasp experience of different objects using deep learning based methods. 



\addtolength{\textheight}{-1cm}   



 
\section*{ACKNOWLEDGMENT}
The authors would like to thank Berkeley AutoLab for providing the 3D printed objects. 

\bibliographystyle{IEEEtran}
\bibliography{references}

\end{document}